\documentclass{article} 
\usepackage{fmocc_arxiv,times}

\usepackage{amsmath,amsfonts,bm}

\def\eqref#1{equation~\ref{#1}}

\def\1{\bm{1}}

\DeclareMathAlphabet{\mathsfit}{\encodingdefault}{\sfdefault}{m}{sl}
\SetMathAlphabet{\mathsfit}{bold}{\encodingdefault}{\sfdefault}{bx}{n}

\usepackage{hyperref}
\usepackage{url}
\usepackage{graphicx} 
\usepackage{adjustbox}

\usepackage{amsmath,amssymb,amsfonts}
\usepackage{textcomp}

\usepackage{booktabs}
\usepackage{multirow}
\usepackage{array}
\usepackage{makecell}
\usepackage{wrapfig}
\usepackage{float}

\usepackage{placeins}
\usepackage{dblfloatfix}
\usepackage{capt-of}

\usepackage{xcolor}
\usepackage{colortbl}

\usepackage[normalem]{ulem}

\providecommand{\NumFont}{\relax}

\newcommand{\best}[1]{{\NumFont\textbf{#1}}}
\newcommand{\second}[1]{{\NumFont\underline{#1}}}

\renewcommand{\arraystretch}{1.15}
\newcolumntype{N}[1]{>{\centering\arraybackslash}m{#1}}

\newcommand{\imp}[2]{%
  \makecell[c]{%
    {\NumFont #1}\\[-0.25ex]
    {\scriptsize\textcolor{gray!70}{(#2)}}%
  }%
}

\newlength{\HdrH}
\newcolumntype{C}[1]{>{\hspace{3pt}\centering\arraybackslash}p{#1}<{\hspace{3pt}}}
\newcolumntype{L}[1]{>{\hspace{3pt}\raggedright\arraybackslash}p{#1}<{\hspace{3pt}}}

\newcommand{\sq}[1]{\textcolor{#1}{\rule{7pt}{7pt}}}

\newcommand{\tablelhead}[1]{%
  \parbox[b][\HdrH][b]{\linewidth}{\raggedright\bfseries #1}%
}

\newcommand{\mhead}[1]{%
  \parbox[b][\HdrH][b]{\linewidth}{\centering\bfseries #1}%
}
\newcommand{\vhead}[3]{%
  \parbox[b][\HdrH][b]{1.05cm}{%
    \centering
    \makebox[0pt][c]{%
      \rotatebox[origin=bl]{90}{%
        \vbox{\offinterlineskip
          \hbox{#2}%
          \kern0.4ex
          \hbox{\scriptsize\textcolor{gray!70}{(#3\%)}}%
        }%
      }%
    }\par
    \vspace{0.55ex}
    \sq{#1}%
  }%
}

\title{FM-Occ: Feature-Diverse Upsampling and Memory-Aware Decoding for 3D Occupancy Prediction}

\author{
\normalfont
Yujia Zhai \quad Guoming Huang \quad
Zhuanghuai Wang \quad Bangquan Xie\textsuperscript{*}
\\[4pt]
\textit{Great Bay University}
\\[3pt]
{\small \textsuperscript{*}Corresponding author}
}

\arxivfinalcopy
\begin{document}

\maketitle

\begin{abstract}
3D semantic occupancy prediction estimates voxel-wise occupancy and semantic labels throughout a scene. Our experiments reveal that conventional sparse transposed-convolution upsampling yields poorly differentiated voxel responses and that successive decoder layers discard previously correct semantic predictions. To address these issues, we present Feature-Diverse Upsampling and
Memory-Aware Decoding for 3D Occupancy Prediction (\textbf{FM-Occ}), comprising two
components. First, 3D Sparse Subvoxel Shuffle (\textbf{3S}) mitigates feature homogenization by constructing candidate child features in the channel dimension, jointly transforming and normalizing them, and then rearranging them onto a finer sparse grid with explicit feature--coordinate correspondence. Second, Entropy-Regulated
Layer-wise Memory (\textbf{ER-LM}) retains semantic information across decoder layers by maintaining a layer-wise memory for each query through gated recurrent updates and using prediction entropy to regulate the historical information propagated to the next layer. During training, a class-weighted entropy loss $\mathcal{L}_{\mathrm{cwe}}$ provides label-guided regularization for matched semantic queries. On SemanticKITTI, 3S increases the spatial coefficient of variation of normalized voxel-wise root-mean-square (RMS) responses from $0.41\%$ to $4.03\%$, while ER-LM reduces semantic degradation and improves error correction across decoder layers. The combined FM-Occ improves mIoU from $12.16\%$ to $12.73\%$ with negligible memory overhead. ER-LM further improves mIoU in the BEV-based DISC framework from $16.78\%$ to $17.17\%$, supporting its applicability across 3D voxel-based and BEV-based architectures.
\end{abstract}

\section{INTRODUCTION}

3D occupancy prediction \citep{cao2022monoscene, wei2023surroundocc, zhang2023occformer}
aims to infer voxel-wise occupancy states and semantic categories in 3D space,
providing a dense and structured scene representation for autonomous driving
and robotic perception. Compared with 3D object detection
\citep{yan2018second, lang2019pointpillars, li2024bevformer}, which outputs
sparse bounding boxes, occupancy prediction explicitly models scene geometry
and spatial continuity. Following sparse occupancy approaches such as SparseOcc
and OPUS \citep{tang2024sparseocc, wang2024opus}, we study an image-based
encoder--decoder architecture that constructs sparse 3D features and
progressively predicts semantic occupancy through feature upsampling and
iterative Transformer decoding.

We further examine these two key stages and identify two overlooked issues: weak spatial differentiation of upsampled voxel features and progressive loss of semantic information during iterative decoding.

\begin{figure*}[!t]
    \centering
    \begin{minipage}[c]{0.46\textwidth}
        \centering
        \includegraphics[width=\linewidth]{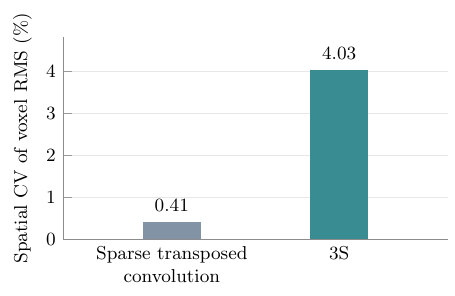}
    \end{minipage}\hfill
    \begin{minipage}[c]{0.51\textwidth}
        \centering
        \includegraphics[width=\linewidth]{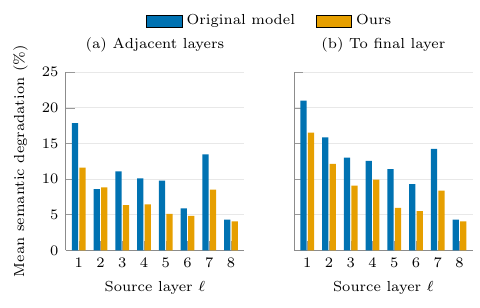}
    \end{minipage}
    \caption{\textbf{Two challenges in occupancy prediction: feature homogenization and fragile semantic consistency.}
    \textbf{Left:} Spatial variation of voxel-wise feature responses after upsampling, measured by spatial CV (higher indicates greater variation).
    \textbf{Right:} Mean class-wise semantic degradation (mD; lower is better) without and with ER-LM + $\mathcal{L}_{\mathrm{cwe}}$, comparing (a) adjacent decoder layers and (b) intermediate layers with the final layer.}
    \label{fig:observations}
\end{figure*}

\textbf{Observation 1: Limited differentiation of upsampled voxel features.} Our analysis shows that sparse transposed convolution followed by group normalization produces voxel features with limited spatial variation in their response magnitudes. This observation suggests that the organization of feature construction and normalization deserves attention alongside the expansion of the voxel grid. We introduce 3D Sparse Subvoxel Shuffle (3S), which constructs candidate child representations in the channel dimension, jointly transforms and normalizes them, and then rearranges them onto the finer sparse grid. Across 50 validation frames, 3S increases the spatial CV of voxel-wise root-mean-square (RMS) responses from $0.41\%$ to $4.03\%$ (Fig.~\ref{fig:observations}, left). 3S also improves both mIoU and IoU.

\textbf{Observation 2: Semantic degradation across decoder layers.} Our analysis reveals substantial semantic degradation during iterative decoding, with a mean class-wise degradation rate of $21.00\%$ from the first to the final decoder layer (Fig.~\ref{fig:observations}, right). To address this issue, we introduce Entropy-Regulated Layer-wise Memory (ER-LM), which maintains a gated memory for each query across decoder layers. Shared recurrent parameters accumulate historical query information, while query-specific classification entropy regulates the memory state propagated to subsequent stages. During training, a complementary class-weighted entropy loss $\mathcal{L}_{\mathrm{cwe}}$ encourages concentrated class predictions on matched positive queries, with greater relative emphasis on rare categories. In the separate decoding experiments, ER-LM trained with $\mathcal{L}_{\mathrm{cwe}}$ improves final mIoU from $12.16\%$ to $12.88\%$ and reduces the mean class-wise semantic degradation rate from $21.00\%$ to $16.50\%$.

Our contributions are summarized as follows:
\begin{itemize}
\item We propose FM-Occ, a 3D semantic occupancy framework that improves feature differentiation during sparse upsampling and semantic preservation across decoder layers.
\item We propose 3S to preserve spatial variation in feature magnitudes during sparse upsampling by reorganizing child-feature construction, normalization, and spatial expansion.
\item We introduce ER-LM to mitigate cross-layer semantic degradation through entropy-regulated memory, complemented by a class-weighted entropy loss during training. ER-LM generalizes across both 3D voxel-based and BEV-based frameworks.
\end{itemize}

\section{RELATED WORK}

\subsection{Representative Approaches for 3D Occupancy Prediction}

Vision-based semantic occupancy methods differ in how they represent and process 3D scenes. Dense voxel-based approaches construct features on regular 3D grids to predict scene geometry and semantics \citep{cao2022monoscene, wang2023openoccupancy, wei2023surroundocc, zhang2023occformer}. BEV-based and tri-perspective representations reduce volumetric processing by projecting or decomposing 3D scene features onto compact planes \citep{hou2024fastocc, xue2024bi, liu2025disentangling, huang2023tri, zuo2023pointocc}. Sparse approaches avoid maintaining features over the entire 3D grid by operating on a limited set of active elements. \citet{tang2024sparseocc} adopts a sparse voxel representation and enhances it through multi-scale feature interaction. \citet{wang2024opus} adopts a different sparse formulation, representing occupancy as a set of occupied locations and semantic classes predicted by learnable queries. These approaches illustrate different choices of scene representation; our work examines feature reconstruction and semantic information retention within sparse voxel and iterative decoding architectures.

\subsection{Feature Upsampling}

2D feature upsampling has been explored through learned kernels, content-aware reassembly, adaptive sampling, and channel-to-space rearrangement \citep{zeiler2010deconvolutional, wang2019carafe, liu2023learning, shi2016real}. These methods operate on dense feature maps with spatial positions implicitly defined by regular tensor grids.
Unlike dense 2D feature maps, sparse 3D representations store only active voxel features together with explicit coordinates. Sparse 3D upsampling must therefore expand both features and coordinates while maintaining their correspondence. 3S addresses this requirement through channel-to-space feature rearrangement and matched coordinate expansion.

\subsection{Information Retention in Occupancy Decoding}

Query-based occupancy methods employ stacked decoder layers to progressively refine semantic predictions \citep{zhang2023occformer, wang2024panoocc, oh20253d}. In these decoders, successive layers repeatedly update the query representations, with information from earlier stages carried forward implicitly through the evolving queries. As a result, previously acquired semantic information may be altered or lost during subsequent refinement, motivating explicit mechanisms for cross-layer information retention.

\section{METHOD}

\begin{figure*}[tb]
    \centering
    \includegraphics[width=\textwidth]{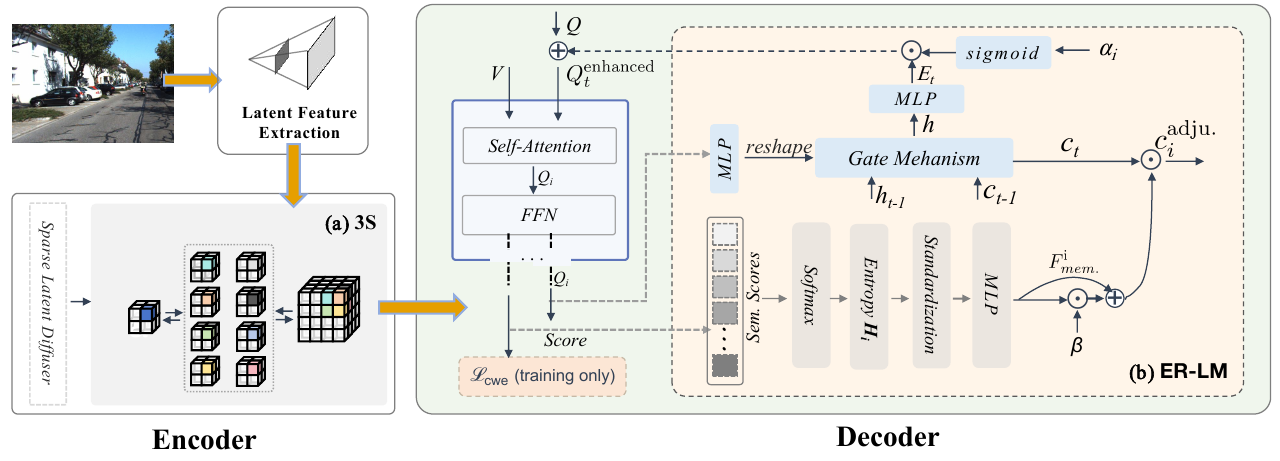}
    \caption{\textbf{Overall framework of FM-Occ.} 3S performs sparse 3D feature upsampling, while ER-LM propagates information across successive decoder stages under entropy-guided regulation. $\mathcal{L}_{\mathrm{cwe}}$ is applied only during training.}
    \label{fig:framework}
\end{figure*}

The overall framework of FM-Occ is illustrated in Fig.~\ref{fig:framework}.
Latent Feature Extraction first maps image features into a sparse 3D representation. 
The Sparse Latent Diffuser then performs scene completion,
followed by a multi-scale sparse feature pyramid. 3S is used to recover
higher-resolution sparse features during multi-scale feature fusion.
Finally, the Transformer decoder predicts semantic occupancy, where ER-LM
propagates information across decoder stages under entropy-guided regulation.
The class-weighted entropy loss $\mathcal{L}_{\mathrm{cwe}}$ is additionally
applied during training.

\subsection{3D Sparse Subvoxel Shuffle}

\begin{figure}[!t]
    \centering
    \includegraphics[width=0.90\linewidth]{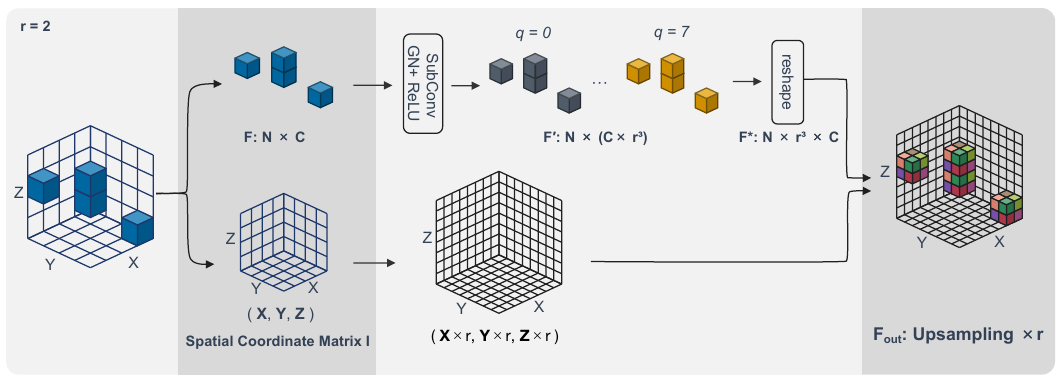}
    \caption{\textbf{3S.}
    The \textbf{upper branch} expands feature channels through a submanifold convolution, applies group normalization and ReLU, and reshapes the features into $r^3$ child representations per parent voxel. The \textbf{lower branch} generates the corresponding fine-grid coordinates. The two branches combine to form the upsampled sparse tensor, illustrated here for $r=2$.}
    \label{fig:3S}
\end{figure}

Sparse feature pyramids rely on upsampling to recover fine-grained voxel
representations from coarse features. Conventional sparse transposed
convolution generates child features while expanding the spatial grid,
coupling feature construction with spatial placement. We introduce
3S, which first generates
$r^3$ candidate child representations in the channel dimension for each
active parent voxel. These representations are jointly normalized and
activated at the coarse resolution before being deterministically
rearranged to their corresponding locations on the finer sparse grid.
The overall process is illustrated in Fig.~\ref{fig:3S}.

\subsubsection{Sparse tensor representation}
Let the input sparse tensor be represented as $\mathcal{S}=(\mathbf{F},\mathbf{I},\mathbf{s})$,
where $\mathbf{F}\in\mathbb{R}^{N\times C_{\mathrm{in}}}$ contains the
features of $N$ active voxels, $\mathbf{I}\in\mathbb{Z}^{N\times4}$ stores
their batch and three-dimensional spatial indices, and
$\mathbf{s}=(S_1,S_2,S_3)$ denotes the spatial shape. The coordinate of the
$n$-th active voxel is written as $\mathbf{I}_n=(b_n,\mathbf{p}_n)$,
where $b_n$ is the batch index and $\mathbf{p}_n=(p_{n,1},p_{n,2},p_{n,3})$
is its three-dimensional spatial coordinate.

\subsubsection{Child-feature construction and joint normalization}
Given an upsampling factor $r$, a submanifold sparse convolution expands each input feature from $C_{\mathrm{in}}$ channels to $r^3C_{\mathrm{out}}$ channels, yielding $\widetilde{\mathbf{F}}=\operatorname{Conv}_{1\times1\times1}(\mathbf{F})$, where $\widetilde{\mathbf{F}}\in\mathbb{R}^{N\times(r^3C_{\mathrm{out}})}$.

Before spatial rearrangement, we apply group normalization and a ReLU activation to the expanded representation, obtaining $\overline{\mathbf{F}}=\operatorname{ReLU}(\operatorname{GN}(\widetilde{\mathbf{F}}))$.

\subsubsection{Channel-to-space rearrangement}
The normalized representation is reshaped as
\begin{equation}
    \overline{\mathbf{F}}
    \in
    \mathbb{R}^{N\times(r^3C_{\mathrm{out}})}
    \longrightarrow
    \mathbf{F}^{\mathrm{sub}}
    \in
    \mathbb{R}^{N\times r^3\times C_{\mathrm{out}}}.
\end{equation}
Let $q\in\{0,\ldots,r^3-1\}$ denote a child index. The exact feature mapping implemented by 3S is
\begin{equation}
    \mathbf{F}^{\mathrm{sub}}_{n,q,c}
    =
    \overline{\mathbf{F}}_{n,\,cr^3+q},
\end{equation}
where $c\in\{0,\ldots,C_{\mathrm{out}}-1\}$. Flattening the parent and child dimensions produces
\begin{equation}
    \mathbf{F}_{\mathrm{out}}
    \in
    \mathbb{R}^{(Nr^3)\times C_{\mathrm{out}}},
    \qquad
    \mathbf{F}_{\mathrm{out}}[nr^3+q,c]
    =
    \mathbf{F}^{\mathrm{sub}}_{n,q,c}.
    \label{eq:3s_feature_mapping}
\end{equation}
Thus, each parent voxel produces $r^3$ location-specific child-feature vectors.

\subsubsection{Coordinate expansion and alignment}
We enumerate the complete set of three-dimensional child offsets as
$\mathcal{D}_r=\{(d_1,d_2,d_3)\mid d_1,d_2,d_3\in\{0,\ldots,r-1\}\}$.
Let $\mathbf{d}_q\in\mathcal{D}_r$ be the offset associated with child index $q$. The output coordinate paired with the feature in Eq.~\eqref{eq:3s_feature_mapping} is
\begin{equation}
    \mathbf{I}_{\mathrm{out}}[nr^3+q]
    =
    \left(
        b_n,\,
        r\mathbf{p}_n+\mathbf{d}_q
    \right).
    \label{eq:3s_coordinate_mapping}
\end{equation}
The output spatial shape is correspondingly updated to
$\mathbf{s}_{\mathrm{out}}=(rS_1,rS_2,rS_3)$.

The resulting sparse tensor is
$\mathcal{S}_{\mathrm{out}}=(\mathbf{F}_{\mathrm{out}},\mathbf{I}_{\mathrm{out}},\mathbf{s}_{\mathrm{out}})$.
Within each sparse pyramid stage, $\mathcal{S}_{\mathrm{out}}$ is
subsequently merged with the existing feature level at the same resolution. 

\subsection{Entropy-Regulated Layer-wise Memory}
\label{sec:er_lm}

Iterative query updates can both correct semantic errors and disrupt previously correct predictions. To support the reuse of semantic evidence across decoding stages, we introduce ER-LM. As illustrated in the right panel of Fig.~\ref{fig:observations}, ER-LM maintains a gated memory alongside the evolving query representations and incorporates its output into subsequent decoder updates. The recurrent module is shared across decoding stages, while classification entropy provides a query-specific signal for modulating memory propagation.

\subsubsection{Gated Memory across Decoding Stages}

Let  $\mathbf{Q}^{(\ell)} \in \mathbb{R}^{B \times N_q \times D}$ denote the query representations after decoder layer $\ell$, where $B$, $N_q$, and $D$ are the batch size, number of queries, and query feature dimension, respectively. The initial learnable queries are denoted by $\mathbf{Q}^{(0)}$. We use an LSTM with hidden dimension $D_h$ to maintain a separate memory state for each query across the $L$ decoding stages.

At the beginning of each forward pass, the hidden and cell states are initialized to zero, $\mathbf{h}^{(0)}=\mathbf{0}$ and
$\widetilde{\mathbf{c}}^{(0)}=\mathbf{0}$, 
where both states have shape $1 \times (B N_q) \times D_h$.

Before decoder layer $\ell$, the incoming queries are reshaped as
\[
\mathbf{X}^{(\ell)}
=\mathcal{R}\!\left(\mathbf{Q}^{(\ell-1)}\right)
\in\mathbb{R}^{1\times(BN_q)\times D}
\]
and passed to the shared LSTM:
\begin{align}
\left(\mathbf{h}^{(\ell)},\mathbf{c}^{(\ell)}\right)
&=
\operatorname{LSTM}
\left(
\mathbf{X}^{(\ell)};
\mathbf{h}^{(\ell-1)},\widetilde{\mathbf{c}}^{(\ell-1)}
\right),
\label{eq:layer_memory_update}
\end{align}
where $\mathcal{R}$ merges the batch and query dimensions, and $\mathcal{R}^{-1}$ restores their original layout. Each LSTM invocation processes one recurrent step, with the hidden and cell states propagated across decoding stages. The standard LSTM gates control the incorporation of current query information and the retention of previous memory. Projecting the hidden state to the query dimension and restoring the
batch--query layout yields the memory feature
$\mathbf{E}^{(\ell)}\in\mathbb{R}^{B\times N_q\times D}$, which is fused
with the incoming queries before decoder layer $\ell$:
\begin{equation}
\mathbf{E}^{(\ell)}
=\mathcal{R}^{-1}\!\left(
\mathbf{h}^{(\ell)}\mathbf{W}_p^{\top}+\mathbf{b}_p
\right).
\end{equation}
\noindent\begin{minipage}[t]{0.49\columnwidth}
\begin{equation}
\widehat{\mathbf{Q}}^{(\ell-1)}
=\mathbf{Q}^{(\ell-1)}+\alpha_{\ell}\mathbf{E}^{(\ell)}.
\end{equation}
\end{minipage}\hfill
\begin{minipage}[t]{0.49\columnwidth}
\vspace{-2pt}
\begin{equation}
\mathbf{Q}^{(\ell)}
=\mathcal{D}_{\ell}\!\left(
\widehat{\mathbf{Q}}^{(\ell-1)},
\mathbf{F}_{s(\ell)}
\right).
\label{eq:memory_query_fusion}
\end{equation}
\end{minipage}

where $\mathbf{W}_p \in \mathbb{R}^{D \times D_h}$ and
$\mathbf{b}_p \in \mathbb{R}^{D}$ are the shared projection matrix and bias, respectively, $\mathcal{D}_{\ell}$ denotes decoder layer $\ell$, and $\mathbf{F}_{s(\ell)}$ is the feature level accessed at that stage. The fusion weight is $\alpha_{\ell}=\sigma(a_{\ell})$; the scalar $a_{\ell}$ is learned separately for each decoder layer and initialized to $0.1$.

\subsubsection{Entropy-Regulated Memory Propagation}

We use classification entropy to modulate the cell state transmitted between decoding stages. Let $\mathbf{S}^{(\ell-1)}$ denote the classification logits obtained from the preceding stage, and let $K$ be the number of classification outputs, including the no-object class. For each query $q$, the class probabilities are $p_{q,k}^{(\ell-1)}=\operatorname{Softmax}(\mathbf{S}_q^{(\ell-1)})_k$. The entropy and its normalized form are
\noindent\begin{minipage}[t]{0.58\columnwidth}
\begin{equation}
H_q^{(\ell-1)}=-\sum_{k=1}^{K}p_{q,k}^{(\ell-1)}
\log\left(p_{q,k}^{(\ell-1)}+\epsilon\right).
\end{equation}
\end{minipage}\hfill
\begin{minipage}[t]{0.40\columnwidth}
\begin{equation}
\overline{H}_q^{(\ell-1)}=\frac{H_q^{(\ell-1)}}{\log K}.
\end{equation}
\end{minipage}
where $\epsilon=10^{-8}$ ensures numerical stability. 

For stages $\ell \geq 2$, a shared multilayer perceptron maps the normalized entropy to a query-specific modulation coefficient:
\begin{equation}
r_q^{(\ell)}
=
\sigma\left(
g\left(\overline{H}_q^{(\ell-1)}\right)
\right).
\end{equation}
where $g(\cdot)$ is a lightweight MLP that maps prediction entropy to a gating coefficient. At the first stage, we use the default coefficient $r_q^{(1)}=0.5$. 

After the LSTM update in Eq.~\eqref{eq:layer_memory_update}, the cell state is modulated as
\begin{equation}
\widetilde{\mathbf{c}}^{(\ell)}
=
\mathbf{c}^{(\ell)}
\odot
\left(
1+\beta\,\mathcal{R}\!\left(\mathbf{r}^{(\ell)}\right)
\right),
\qquad
\beta=0.3.
\label{eq:entropy_cell_modulation}
\end{equation}
The coefficient $\beta=0.3$ bounds the modulation gain between $1$ and $1.3$.
The adjusted cell state $\widetilde{\mathbf{c}}^{(\ell)}$ is passed to the next recurrent step, while the current hidden output $\mathbf{h}^{(\ell)}$ is used for query fusion. 

\subsubsection{Class-Weighted Entropy Regularization}
\label{sec:cwe}

Because the prediction entropy used by ER-LM is not directly supervised,
we introduce $\mathcal{L}_{\mathrm{cwe}}$ to explicitly regularize it on
matched positive queries. Ground-truth assignments determine the positive
queries and their class weights. This entropy supervision also reduces
uncertainty in the predicted class distributions, complementing the
classification loss.

For prediction stage $\ell$, let $\mathcal{Q}_{+}^{(\ell)}$ denote the queries matched to ground-truth semantic masks, excluding those assigned to the no-object class; let $y_q^{(\ell)}$ denote their assigned labels, and $w_{y_q^{(\ell)}}$ the corresponding class weights used by the classification loss. The stage-wise loss is defined as
\begin{equation}
\mathcal{L}_{\mathrm{cwe}}^{(\ell)}
=
\Bigl(\sum\nolimits_{q \in \mathcal{Q}_{+}^{(\ell)}}
w_{y_q^{(\ell)}} H_q^{(\ell)}
\Bigr)\Big/\Bigl(
\sum\nolimits_{q \in \mathcal{Q}_{+}^{(\ell)}}
w_{y_q^{(\ell)}}
\Bigr).
\label{eq:cwe}
\end{equation}

We apply this regularizer to the initial query predictions and the outputs of all decoder layers; the overall training objective is
\begin{equation}
\mathcal{L}_{\mathrm{total}}
=\mathcal{L}_{\mathrm{base}}
+\lambda_{\mathrm{cwe}}\mathcal{L}_{\mathrm{cwe}},
\qquad
\mathcal{L}_{\mathrm{cwe}}
=\sum_{\ell=0}^{L}\mathcal{L}_{\mathrm{cwe}}^{(\ell)}.
\end{equation}
\par\noindent where $\ell=0$ denotes the initial prediction stage, $L$ is the number of decoder layers, and $\mathcal{L}_{\mathrm{base}}$ comprises the original classification, mask, Dice, and voxel-level losses. We set $\lambda_{\mathrm{cwe}}=1.0$ in our experiments.

\section{EXPERIMENTAL RESULTS}

\subsection{Experimental Setup}
We evaluate FM-Occ on the SemanticKITTI \citep{behley2019semantickitti} dataset. SemanticKITTI is a large-scale dataset for 3D semantic occupancy prediction in autonomous driving scenarios, based on the raw KITTI sequences. The dataset contains 21 classes, including 19 semantic categories, one free-space class, and one unknown class. We train FM-Occ for 30 epochs on four NVIDIA RTX 4090 GPUs using AdamW with an initial learning rate of $1\times10^{-4}$ and a total batch size of 4. We adopt a step learning-rate schedule. 

We adopt commonly used evaluation metrics for 3D occupancy prediction, including voxel-wise Intersection over Union (IoU) and mean Intersection over Union (mIoU). To provide a more systematic analysis of model performance across different semantic structures, we further propose category-grouped evaluation metrics, namely ComplexIoU, TailIoU, and RegularIoU. Detailed definitions and the category grouping of these metrics are provided in Appendix~\ref{app:metrics}.

\subsection{Comparison with state-of-the-art}

Table~\ref{tab:ssc_final_fixed} compares FM-Occ with recent semantic scene completion methods on SemanticKITTI. FM-Occ achieves an IoU of $36.18\%$ and an mIoU of $12.73\%$. To further evaluate the transferability of ER-LM, we apply it from the sparse 3D voxel-based FM-Occ to the BEV-based DISC framework, improving DISC from $16.78\%$ to $17.17\%$ mIoU and from $45.54\%$ to $45.67\%$ IoU. The resulting $17.17\%$ mIoU is comparable to MRA ($17.14\%$) and higher than SGFormer ($16.68\%$), demonstrating competitive performance against recent methods and the transferability of ER-LM across different scene representations.

\definecolor{cRoad}{RGB}{255,0,200}
\definecolor{cSidewalk}{RGB}{60,0,80}
\definecolor{cParking}{RGB}{255,150,255}
\definecolor{cOtherGround}{RGB}{160,0,80}
\definecolor{cBuilding}{RGB}{255,190,0}
\definecolor{cCar}{RGB}{90,140,255}
\definecolor{cTruck}{RGB}{70,40,180}
\definecolor{cBicycle}{RGB}{80,230,255}
\definecolor{cMotorcycle}{RGB}{25,55,160}
\definecolor{cOtherVeh}{RGB}{90,70,255}
\definecolor{cVegetation}{RGB}{30,160,30}
\definecolor{cTrunk}{RGB}{120,60,0}
\definecolor{cTerrain}{RGB}{160,255,90}
\definecolor{cPerson}{RGB}{255,30,30}
\definecolor{cBicyclist}{RGB}{255,60,230}
\definecolor{cMotorcyclist}{RGB}{150,30,90}
\definecolor{cFence}{RGB}{255,120,40}
\definecolor{cPole}{RGB}{255,235,120}
\definecolor{cTrafSign}{RGB}{255,0,0}
\newcommand{\rotcls}[3]{\rotatebox{90}{\sq{#1}\,#2 (#3)}}

\newcommand{\na}{--}

\begin{table*}[htbp]
\caption{Quantitative comparison on the SemanticKITTI validation set}
\label{tab:ssc_final_fixed}
\centering
\small
\setlength{\tabcolsep}{0pt}
\renewcommand{\arraystretch}{1.20}

\begin{adjustbox}{width=\textwidth}
    
\begin{tabular}{L{4.5cm}|*{2}{C{1.05cm}}|*{19}{C{1.05cm}}}

\toprule
\tablelhead{Method} &
\mhead{IoU} & \mhead{\mbox{mIoU}} &

\vhead{cRoad}{road}{15.30} &
\vhead{cSidewalk}{sidewalk}{11.13} &
\vhead{cParking}{parking}{1.12} &
\vhead{cOtherGround}{other-grnd.}{0.56} &
\vhead{cBuilding}{building}{14.10} &
\vhead{cCar}{car}{3.92} &
\vhead{cTruck}{truck}{0.16} &
\vhead{cBicycle}{bicycle}{0.03} &
\vhead{cMotorcycle}{motorcycle}{0.03} &
\vhead{cOtherVeh}{other-veh.}{0.20} &
\vhead{cVegetation}{vegetation}{39.30} &
\vhead{cTrunk}{trunk}{0.51} &
\vhead{cTerrain}{terrain}{9.17} &
\vhead{cPerson}{person}{0.07} &
\vhead{cBicyclist}{bicyclist}{0.07} &
\vhead{cMotorcyclist}{motorcyclist}{0.05} &
\vhead{cFence}{fence}{3.90} &
\vhead{cPole}{pole}{0.29} &
\vhead{cTrafSign}{traf.-sign}{0.08} \\
\midrule

TPVFormer†{\scriptsize\citep{huang2023tri}}  &
35.61 & 11.36 &
56.50 & 25.87 & 20.60 & 0.85 & 13.88 & 23.81 & 8.08 & 0.36 & 0.05 & 4.35 &
16.92 & 2.26 & 30.38 & 0.51 & 0.89 & 0.00 & 5.94 & 3.14 & 1.52 \\

MonoScene†{\scriptsize\citep{cao2022monoscene}} &
36.86 & 11.08 &
56.52 & 26.72 & 14.27 & 0.46 & 14.09 & 23.26 & 6.98 & 0.61 & 0.45 & 1.48 &
17.89 & 2.81 & 29.64 & 1.86 & 1.20 & 0.00 & 5.84 & 4.14 & 2.25 \\

OccFormer{\scriptsize\citep{zhang2023occformer}} &
36.50 & 13.46 &
58.85 & 26.88 & 19.61 & 0.31 & 14.40 & 25.09 & \best{25.53} & 0.81 & 1.19 & 8.52 &
19.63 & 3.93 & 32.62 & 2.78 & 2.82 & 0.00 & 5.61 & 4.26 & 2.86 \\

VoxFormer*{\scriptsize\citep{li2023voxformer}} &
43.92 & 13.12 &
54.44 & 27.05 & 16.97 & 0.28 & 19.64 & 26.78 & 7.66 & 1.50 & 0.41 & 5.68 &
26.12 & 6.33 & 32.80 & 1.15 & 0.26 & 0.00 & 7.93 & 8.94 & 5.41 \\

Symphonies{\scriptsize\citep{jiang2024symphonize}} &
41.92 & 14.89 &
56.37 & 27.58 & 15.28 & 0.95 & 21.64 & 28.68 & \second{20.44} & 2.54 & 2.82 & \best{13.89} &
25.72 & 6.60 & 30.87 & 3.52 & 2.24 & 0.00 & 8.40 & 9.57 & 5.76 \\

SGFormer{\scriptsize\citep{guo2025sgformer}}&
45.01 & 16.68 &
60.85 & 31.53 & 19.32 & \best{2.14} & \best{26.05} & 32.17 & 20.30 & \second{2.95} & \second{3.10} & 11.55 &
27.11 & \second{8.28} & 38.47 & 3.66 & 1.49 & 0.00 & 9.28 & \second{11.58} & \second{7.22} \\

MRA{\scriptsize\citep{yang2026multi}} &
\best{47.28} & \second{17.14} & 61.92 & \second{33.97} & \best{27.35} & 0.06 & \second{25.89} & \best{34.63} & 20.12 & 0.53 & 1.69 & 11.06 & \best{30.08} & \best{9.54} & \second{38.51} & 0.54 & 0.16 & 0.00 & \best{10.13} & \best{13.15} & 6.27 \\

SparseOcc*{\scriptsize\citep{tang2024sparseocc}} &
35.21 & 12.15 &
57.43 & 27.59 & 21.18 & 0.00 & 14.97 & 25.32 & 1.62 & 0.35 & 0.92 & 11.51 &
18.25 & 3.29 & 31.17 & 3.82 & 0.40 & 0.00 & 6.41 & 4.06 & 2.59 \\

DISC*{\scriptsize\citep{liu2025disentangling}} &
45.54 & 16.78 &
\best{63.65} & \best{34.35} & 24.46 & 1.56 & 23.42 & 32.95 & 15.55 & \best{3.04} & 3.06 & 8.06 &
27.41 & 7.66 & \best{39.01} & \second{3.82} & \best{3.11} & 0.00 & \second{9.73} & 11.08 & 6.88 \\

\rowcolor{gray!15}
FM-Occ &
36.18 &
12.73 &
59.00 &
29.03 &
20.55 &
0.62 &
15.88 &
25.56 &
7.54 &
0.48 &
0.73 &
10.72 &
18.84 &
3.48 &
31.83 &
2.66 &
1.69 &
0.00 &
6.12 &
4.31 &
2.77 \\

\rowcolor{gray!15}
DISC + ER-LM &
\imp{\second{45.67}}{+0.13} &
\imp{\best{17.17}}{+0.39} &
\imp{\second{62.83}}{-0.82} &
\imp{33.96}{-0.39} &
\imp{\second{24.66}}{+0.20} &
\imp{\second{2.04}}{+0.48} &
\imp{23.66}{+0.24} &
\imp{\second{33.00}}{+0.05} &
\imp{19.98}{+4.43} &
\imp{2.93}{-0.11} &
\imp{\best{3.19}}{+0.13} &
\imp{\second{11.65}}{+3.59} &
\imp{\second{27.61}}{+0.20} &
\imp{7.90}{+0.24} &
\imp{37.71}{-1.30} &
\imp{\best{4.02}}{+0.20} &
\imp{\second{2.95}}{-0.16} &
\imp{\best{0.01}}{+0.01} &
\imp{9.56}{-0.17} &
\imp{10.97}{-0.11} &
\imp{\best{7.56}}{+0.68} \\

\arrayrulecolor{black}
\bottomrule
\end{tabular}
\end{adjustbox}
\vspace{2pt}
\footnotesize
\raggedright
\noindent \textbf{Note:} * denotes results reproduced by our implementation, and † denotes results taken from \citet{li2023voxformer}. For DISC + ER-LM, values in parentheses indicate differences from the reproduced DISC baseline. The best results are highlighted in \textbf{bold}, and the second-best results are underlined.

\end{table*}

\subsection{Ablation Studies}

\begin{wraptable}[8]{r}{0.51\columnwidth}
    \vspace{-13pt}
    \centering
    \caption{Ablation studies of 3S, ER-LM, and $\mathcal{L}_{\mathrm{cwe}}$}
    \label{tab:ablation}
    \footnotesize
    \setlength{\tabcolsep}{3.8pt}
    \renewcommand{\arraystretch}{1.14}

    \begin{adjustbox}{max width=\linewidth}
    \begin{tabular}{@{}ccc|cccccc@{}}
        \toprule
        \textbf{3S} &
        \textbf{ER-LM} &
        \(\boldsymbol{\mathcal{L}}_{\mathrm{cwe}}\) &
        \textbf{IoU} &
        \textbf{mIoU} &
        \textbf{Complex}  & 
        \textbf{Tail} & 
        \textbf{Regular} &
        \textbf{Mem.} \\
        \midrule

        & & &
        36.13 & 12.16 & 23.00 & 1.36 & 10.32 & 13.05 \\

        \checkmark & & &
        \textbf{36.62} & 12.42 & 23.14 & 1.53 & 10.80 & 13.06 \\

        & \checkmark & &
        36.11 & 12.67 & 23.07 & \textbf{1.81} & 11.39 & 13.06 \\

        & \checkmark & \checkmark &
        35.85 & \textbf{12.88} & 23.31 & 1.55 & \textbf{12.04} & 13.06 \\

        \checkmark & \checkmark & \checkmark &
        36.18 & 12.73 & \textbf{23.34} & 1.39 & 11.69 & 13.09 \\

        \bottomrule
    \end{tabular}
    \end{adjustbox}
\end{wraptable}

Table~\ref{tab:ablation} compares the proposed components and their combinations.
3S improves IoU and mIoU by $0.49\%$ and $0.26\%$, respectively.
ER-LM primarily improves semantic prediction, increasing mIoU by $0.51\%$
while maintaining comparable occupancy IoU. Adding
$\mathcal{L}_{\mathrm{cwe}}$ further increases mIoU to $12.88\%$.
The full configuration improves all five accuracy metrics over the baseline,
although the gains of the individual components are not strictly additive.
These improvements introduce negligible memory overhead, with measured GPU
memory increasing from $13.05$ to $13.09$ GB.

\begin{table}[!ht]
\begin{minipage}[c]{0.45\columnwidth}
Table~\ref{tab:five_seed} further evaluates robustness across five random seeds. Both 3S and ER-LM retain their principal improvements across repeated runs. 3S shows stable semantic performance, while ER-LM provides a consistent mIoU gain and reduces the variation on regular categories. These results indicate that the observed improvements are not dependent on a single favorable training run.
\end{minipage}
\hfill
\begin{minipage}[c]{0.53\columnwidth}
\vspace{-4pt}
\centering
\caption{Robustness evaluation across five random seeds. Results are reported as mean $\pm$ standard deviation}
\label{tab:five_seed}
\footnotesize
\setlength{\tabcolsep}{3.0pt}
\renewcommand{\arraystretch}{1.15}

\begin{adjustbox}{max width=\linewidth}
\begin{tabular}{@{}lccccc@{}}
\toprule
\textbf{Method} & \textbf{IoU} & \textbf{mIoU} & \textbf{Complex}  & \textbf{Tail} & \textbf{Regular} \\
\midrule

{-{}-} &
$35.91${\scriptsize\textcolor{gray}{$\pm 0.16$}} &
$12.34${\scriptsize\textcolor{gray}{$\pm 0.21$}} &
$22.96${\scriptsize\textcolor{gray}{$\pm 0.34$}} &
$1.29${\scriptsize\textcolor{gray}{$\pm 0.16$}} &
$11.31${\scriptsize\textcolor{gray}{$\pm 0.95$}} \\

+ 3S &
$\mathbf{36.18}${\scriptsize\textcolor{gray}{$\pm 0.32$}} &
$12.44${\scriptsize\textcolor{gray}{$\pm 0.12$}} &
$\mathbf{23.15}${\scriptsize\textcolor{gray}{$\pm 0.36$}} &
$\mathbf{1.49}${\scriptsize\textcolor{gray}{$\pm 0.27$}} &
$10.89${\scriptsize\textcolor{gray}{$\pm 0.46$}} \\

+ ER-LM &
$35.91${\scriptsize\textcolor{gray}{$\pm 0.26$}} &
$\mathbf{12.56}${\scriptsize\textcolor{gray}{$\pm 0.25$}} &
$22.78${\scriptsize\textcolor{gray}{$\pm 0.51$}} &
$1.32${\scriptsize\textcolor{gray}{$\pm 0.15$}} &
$\mathbf{11.87}${\scriptsize\textcolor{gray}{$\pm 0.39$}} \\

\bottomrule
\end{tabular}
\end{adjustbox}

\par\vspace{2pt}
\scriptsize\raggedright
\textit{Note.} All configurations were re-run under the same setting, which differs slightly from Table~\ref{tab:ablation}.
\end{minipage}
\end{table}
\FloatBarrier

\subsection{Analysis of 3S}

\subsubsection{Comparison of Upsampling Feature Responses}

\begin{table}[!ht]
\begin{minipage}[c]{0.66\columnwidth}
\vspace{0pt}

We further examine the spatial variation of voxel-wise feature responses
produced by different upsampling schemes. As shown in Table~\ref{tab:spatial_variation}, 3S achieves a mean spatial
coefficient of variation (CV) of $4.03\%$, compared with $0.41\%$ for
sparse transposed convolution, corresponding to a $9.83\times$ increase.
Figure~\ref{fig:3s_response} further shows clearer response separation
between the car and its surrounding background in region A and between
vegetation and the ground in regions B and C. The definition and measurement protocol of the feature-response variation
metric, together with details of the feature-response visualization and
analysis, are provided in Appendix~\ref{app:feature_response}.
\end{minipage}
\hfill
\begin{minipage}[c]{0.31\columnwidth}
\vspace{-9pt}
\centering
\caption{Comparison of spatial variation in voxel-wise RMS responses}
\label{tab:spatial_variation}
\footnotesize
\setlength{\tabcolsep}{3pt}
\renewcommand{\arraystretch}{1.12}
\begin{tabular}{@{}lcc@{}}
\toprule
\textbf{Method} & \textbf{CV (\%)} & \textbf{Ratio} \\
\midrule
Transposed conv. & 0.41 & $1.00\times$ \\
3S & \textbf{4.03} & \textbf{$9.83\times$} \\
\bottomrule
\end{tabular}
\end{minipage}
\end{table}

\begin{figure}[!ht]
    \centering
    \setlength{\abovecaptionskip}{0pt}
    \vspace{-8pt}
    \includegraphics[
    width=\linewidth,
    trim=1 3 1 3,
    clip
]{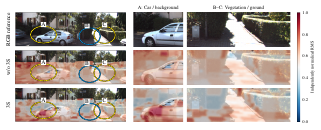}
    \caption{\textbf{Feature-response comparison after upsampling.}
    Full images are shown on the left, with enlarged regions on the right.
    Each response map is independently min--max normalized for visualization.}
    \label{fig:3s_response}
    \vspace{-3pt}
\end{figure}

\subsubsection{Effect of Normalization and Activation Placement}
\label{sec:upsampling_controls}

We examine two alternative explanations for the upsampling results: whether the transposed-convolution reference is disadvantaged by the absence of ReLU after normalization, and whether moving normalization after spatial rearrangement improves 3S. As shown in Table~\ref{tab:upsampling_controls}, adding ReLU to transposed convolution does not improve mIoU, suggesting that ReLU alone does not account for the gains of 3S.
Applying GN after rearrangement gives similar mIoU to the original 3S, suggesting that normalization order has a limited effect on this metric in our experiments.

\subsubsection{Subvoxel Feature Allocation Variants}

As shown in Table~\ref{tab:allocation_variants}, permute--reshape interprets
the expanded channels as $r^3\times C_{\mathrm{out}}$, whereas 3S uses
$C_{\mathrm{out}}\times r^3$. Its mIoU improvement from $12.16\%$ to
$12.32\%$ supports channel expansion and rearrangement, while the $12.42\%$
achieved by 3S validates the proposed feature allocation. Softmax improves mIoU and RegularIoU, but its sum-to-one competition reduces
occupancy IoU. Sigmoid improves ComplexIoU and RegularIoU but lowers TailIoU,
possibly because it favors dominant classes under imbalanced supervision.
We therefore retain deterministic rearrangement for its balanced performance.
Both variants are defined in Appendix~\ref{app:allocation_variants}.

\begin{table}[htbp]
\centering
\begin{minipage}[t]{0.49\textwidth}
    \vspace{0pt}
    \centering
    \begin{minipage}[t][2.7\baselineskip][t]{\linewidth}
    \caption{Effect of activation and normalization placement}
    \label{tab:upsampling_controls}
    \end{minipage}\par
    \scriptsize
    \setlength{\tabcolsep}{1.5pt}
    \renewcommand{\arraystretch}{1.14}
    \begin{adjustbox}{max width=\linewidth}
    \begin{tabular}{@{}lccccc@{}}
        \toprule
        \textbf{Method} & \textbf{IoU} & \textbf{mIoU} & \textbf{Complex}  & \textbf{Tail} & \textbf{Regular} \\
        \midrule
        Transposed conv. + GN & 36.13 & 12.16 & 23.00 & 1.36 & 10.32 \\
        \quad + ReLU & 36.17 & 12.10 & 22.95 & 1.34 & 10.18 \\
        Shuffle $\to$ GN $\to$ ReLU & 36.22 & 12.37 & 23.10 & \textbf{2.02} & 10.20 \\
        3S (GN $\to$ ReLU $\to$ shuffle) & \textbf{36.62} & \textbf{12.42} & \textbf{23.14} & 1.53 & \textbf{10.80} \\
        \bottomrule
    \end{tabular}
    \end{adjustbox}
\end{minipage}\hfill
\begin{minipage}[t]{0.49\textwidth}
    \vspace{0pt}
    \centering
    \begin{minipage}[t][2.2\baselineskip][t]{\linewidth}
    \caption{Subvoxel feature allocation}
    \label{tab:allocation_variants}
    \end{minipage}\par
    \scriptsize
    \setlength{\tabcolsep}{1.5pt}
    \renewcommand{\arraystretch}{1.14}
    \begin{adjustbox}{max width=\linewidth}
    \begin{tabular}{@{}lccccc@{}}
        \toprule
        \textbf{Method} & \textbf{IoU} & \textbf{mIoU} & \textbf{Complex}  & \textbf{Tail} & \textbf{Regular} \\
        \midrule
        \mbox{-{}-} & 36.13 & 12.16 & 23.00 & 1.36 & 10.32 \\
        Permute--reshape & 36.17 & 12.32 & 22.91 & 1.49 & 10.81 \\
         3S & \textbf{36.62} & 12.42 & 23.14 & \textbf{1.53} & 10.80 \\
        \quad + Softmax allocation & 35.92 & \textbf{12.55} & 23.18 & 1.36 & 11.33 \\
        \quad + Sigmoid allocation & 36.35 & 12.51 & \textbf{23.36} & 1.02 & \textbf{11.34} \\
        \bottomrule
    \end{tabular}
    \end{adjustbox}
\end{minipage}
\end{table}

\FloatBarrier

\subsection{Analysis of ER-LM}

\begin{table}[htbp]
\begin{minipage}[t]{0.55\columnwidth}
\vspace{-6pt}
As shown in Table~\ref{tab:er_lm_analysis}, the results of EMA demonstrate the value of historical information, but simple exponential smoothing yields only limited semantic gains. ER-LM achieves a larger mIoU improvement through learnable memory and forgetting, with more balanced improvements across category groups. We further evaluate applying LayerNorm to the LSTM output before projection, with the corresponding ablation results provided in Appendix~\ref{app:er_lm_layernorm}.
\end{minipage}
\hfill
\begin{minipage}[t]{0.42\columnwidth}
\vspace{-10pt}
    \centering
    \caption{Analysis of historical information aggregation}
    \label{tab:er_lm_analysis}
    \scriptsize
    \setlength{\tabcolsep}{3.0pt}
    \renewcommand{\arraystretch}{1.14}
    \begin{adjustbox}{width=\linewidth}
    \begin{tabular}{@{}lccccc@{}}
        \toprule
        \textbf{Method} & \textbf{IoU} & \textbf{mIoU} & \textbf{Complex}  & \textbf{Tail} & \textbf{Regular}\\
        \midrule
        \mbox{-{}-} & 35.78 & 12.16 & 22.84 & 1.15 & 11.33 \\
        EMA & \textbf{36.19} & 12.37 & 22.95 & \textbf{1.63} & 10.77 \\
        ER-LM & 36.09 & \textbf{12.53} & \textbf{22.96} & 1.32 & \textbf{11.57} \\
        \bottomrule
    \end{tabular}
    \end{adjustbox}
    \par\vspace{2pt}
    \scriptsize\raggedright
    \textit{Note.} All configurations were re-run under the same setting, which differs slightly from Table~\ref{tab:ablation}.
\end{minipage}
\end{table}

\label{sec:cross_layer_analysis}

We use mean class-wise degradation (mD) and correction (mR) rates to
evaluate semantic transitions across decoder layers. As shown in
Table~\ref{tab:cross_layer_transitions}, ER-LM +
$\mathcal{L}_{\mathrm{cwe}}$ reduces degradation and improves correction
in most adjacent-layer transitions and all intermediate-to-final
comparisons. From the first to the final layer, mD decreases from
$21.00\%$ to $16.50\%$, while mR increases from $1.16\%$ to $1.43\%$.
Detailed definitions and evaluation protocols are provided in
Appendix~\ref{app:semantic_transition_metrics}.

\begin{table}[!ht]
\centering
\caption{Class-balanced semantic transition rates (\%). Lower mD and higher mR are better}
\label{tab:cross_layer_transitions}
\scriptsize
\setlength{\tabcolsep}{2.2pt}

\begin{tabular*}{\columnwidth}{@{\extracolsep{\fill}}lrrrr@{\hspace{7pt}}lrrrr@{}}
\toprule

& \multicolumn{2}{c}{mD $\downarrow$} & \multicolumn{2}{c}{mR $\uparrow$} &
& \multicolumn{2}{c}{mD $\downarrow$} & \multicolumn{2}{c}{mR $\uparrow$} \\
\cmidrule(lr){2-3}\cmidrule(lr){4-5}
\cmidrule(lr){7-8}\cmidrule(lr){9-10}

Transition & \mbox{-{}-} & Ours & \mbox{-{}-} & Ours &
Transition & \mbox{-{}-} & Ours & \mbox{-{}-} & Ours \\
\midrule

$1\to2$ & 17.86 & \textbf{11.60} & \textbf{0.99} & 0.88 &
$1\to9$ & 21.00 & \textbf{16.50} & 1.16 & \textbf{1.43} \\

$2\to3$ & \textbf{8.60} & 8.84 & 0.64 & \textbf{0.99} &
$2\to9$ & 15.85 & \textbf{12.12} & 0.84 & \textbf{1.50} \\

$3\to4$ & 11.06 & \textbf{6.34} & 0.57 & \textbf{0.72} &
$3\to9$ & 13.01 & \textbf{9.08} & 0.73 & \textbf{1.33} \\

$4\to5$ & 10.10 & \textbf{6.45} & 0.55 & \textbf{0.99} &
$4\to9$ & 12.56 & \textbf{9.92} & 0.77 & \textbf{1.39} \\

$5\to6$ & 9.77 & \textbf{5.11} & 0.51 & \textbf{0.73} &
$5\to9$ & 11.40 & \textbf{5.95} & 0.71 & \textbf{1.03} \\

$6\to7$ & 5.89 & \textbf{4.84} & 0.60 & \textbf{0.87} &
$6\to9$ & 9.29 & \textbf{5.50} & 0.65 & \textbf{0.95} \\

$7\to8$ & 13.46 & \textbf{8.50} & 0.45 & \textbf{0.92} &
$7\to9$ & 14.22 & \textbf{8.39} & 0.56 & \textbf{0.97} \\

$8\to9$ & 4.31 & \textbf{4.08} & 0.38 & \textbf{0.55} &
$8\to9$ & 4.31 & \textbf{4.08} & 0.38 & \textbf{0.55} \\

\bottomrule
\end{tabular*}

\par\vspace{3pt}
\begin{minipage}{\columnwidth}
\scriptsize\raggedright
\textit{Note.} \mbox{-{}-} denotes the model without ER-LM or
$\mathcal{L}_{\mathrm{cwe}}$; Ours denotes ER-LM +
$\mathcal{L}_{\mathrm{cwe}}$. Bold indicates the better value within each layer pair.
\end{minipage}
\end{table}

\section{CONCLUSION}

We presented FM-Occ for sparse 3D semantic occupancy prediction, addressing feature homogenization during upsampling and semantic degradation across decoder layers. The proposed 3S constructs and jointly normalizes child features before spatial rearrangement, preserving greater variation in voxel feature responses. ER-LM retains historical query information through gated recurrent memory and entropy-guided state modulation, complemented by the class-weighted entropy loss $\mathcal{L}_{\mathrm{cwe}}$. Experiments on SemanticKITTI demonstrate improved occupancy and semantic prediction, while the cross-layer analysis shows better preservation of correct semantic predictions and correction of existing errors on average across classes.

\bibliography{fmocc_arxiv}
\bibliographystyle{fmocc_arxiv}

\appendix
\section{Appendix}

\subsection{Definition of Category-Grouped Evaluation Metrics}
\label{app:metrics}
These metrics are computed as the average IoU over predefined category groups, enabling a more fine-grained evaluation beyond overall performance. Formally, given a category group $\mathcal{G}$, the group-wise IoU is defined as:
\begin{equation}
\mathrm{IoU}_{\mathcal{G}} = \frac{1}{|\mathcal{G}|} \sum_{c \in \mathcal{G}} \mathrm{IoU}_c,
\end{equation}
where $\mathrm{IoU}_c$ denotes the IoU of category $c$.

We divide the 19 non-empty semantic categories into three groups based on structural complexity, boundary ambiguity, and sample imbalance, aiming to enable a more reasonable evaluation and a more accurate analysis of the training results.

\textbf{ComplexIoU} measures performance on categories that typically occupy extended spatial regions, exhibit irregular structures, or have ambiguous semantic boundaries. This group includes road {\scriptsize(15.30\%)}, sidewalk {\scriptsize(11.13\%)}, parking {\scriptsize(1.12\%)}, other-ground {\scriptsize(0.56\%)}, terrain {\scriptsize(9.17\%)}, vegetation {\scriptsize(39.30\%)}, and trunk {\scriptsize(0.51\%)}, with the percentages indicating the proportion of each category in the dataset.

\textbf{TailIoU} focuses on long-tail categories with limited samples (all below 0.1\%), where reliable feature learning is challenging. This group includes bicyclist {\scriptsize(0.07\%)}, person {\scriptsize(0.07\%)}, motorcyclist {\scriptsize(0.05\%)}, bicycle {\scriptsize(0.03\%)}, motorcycle {\scriptsize(0.03\%)}, and traffic-sign {\scriptsize(0.08\%)}.

\textbf{RegularIoU} evaluates categories with relatively regular geometric structures and well-defined shapes. This group includes car {\scriptsize(3.92\%)}, building {\scriptsize(14.10\%)}, truck {\scriptsize(0.16\%)}, fence {\scriptsize(3.90\%)}, pole {\scriptsize(0.29\%)}, and other-vehicle {\scriptsize(0.20\%)}.

\subsection{Additional Analysis of Upsampling Feature Responses}
\label{app:feature_response}

\subsubsection{Definition of the Feature-Response Variation Metric}
The response measure and spatial CV reported in the main paper are defined as
follows. For a frame containing $N$ active voxels, let
$\mathbf{x}_n\in\mathbb{R}^{C}$ denote the feature vector of the $n$-th
active voxel, where $C$ is the number of feature channels. Its root-mean-square (RMS) 
response is defined as
\begin{equation}
    R_n
    =
    R(\mathbf{x}_n)
    =
    \sqrt{
        \frac{1}{C}
        \sum_{j=1}^{C}x_{n,j}^{2}
    },
\end{equation}
where $j$ indexes the feature channels.

Let $\mu_R$ and $\sigma_R$ denote the mean and standard deviation,
respectively, of the voxel responses $\{R_n\}_{n=1}^{N}$ within the frame.
The spatial coefficient of variation is then defined as
\begin{equation}
    \mathrm{CV}
    =
    \frac{\sigma_R}{\mu_R}\times 100\%.
\end{equation}

The measurements are taken after group normalization at the second sparse
upsampling stage over 50 uniformly sampled SemanticKITTI validation frames,
and the reported CV is averaged over these frames. The ratio in
Table~\ref{tab:spatial_variation} is computed as the mean CV of each method
divided by that of sparse transposed convolution.

\subsubsection{Feature-Response Visualization Setup and Analysis}

Fig.~\ref{fig:3s_response} compares the second-stage upsampling outputs before
feature fusion. Both outputs use the same grid resolution and projected sample
locations. Region A covers a car and its surrounding area, while regions B and C
correspond to vegetation and ground areas, respectively. Compared with sparse
transposed convolution, 3S exhibits clearer local variation in feature-response
magnitudes within these regions. Each response map is
independently min--max normalized; the visualization therefore compares
relative spatial patterns rather than absolute response magnitudes or semantic
prediction accuracy.

\subsection{Definitions of Softmax- and Sigmoid-Based Feature Allocation}
\label{app:allocation_variants}

Let $\mathbf{H}_{n}\in\mathbb{R}^{r^3\times C_{\mathrm{out}}}$ denote the
candidate child representations of the $n$-th parent voxel. We apply mean
pooling along the child dimension and feed the resulting feature into a
two-layer MLP. The MLP produces
$\mathbf{a}_{n}\in\mathbb{R}^{r^3}$, where $a_{n,q}$ is the allocation
score predicted for the $q$-th candidate child voxel. We convert these
scores into weights using either Softmax or Sigmoid.

The first variant applies Softmax across the $r^3$ child positions:
\begin{equation}
g_{n,q}^{\mathrm{softmax}}
=
\frac{\exp(a_{n,q})}
{\sum_{j=1}^{r^3}\exp(a_{n,j})},
\qquad q=1,\ldots,r^3.
\label{eq:softmax_allocation}
\end{equation}
The resulting weights sum to one, introducing competition among the
candidate child positions.

The second variant applies Sigmoid independently to each allocation score:
\begin{equation}
g_{n,q}^{\mathrm{sigmoid}}
=
\frac{1}{1+\exp(-a_{n,q})},
\qquad q=1,\ldots,r^3.
\label{eq:sigmoid_allocation}
\end{equation}
Unlike Softmax, Sigmoid does not impose a sum-to-one constraint, allowing
multiple child positions to retain relatively large responses
simultaneously. In both variants, each weight $g_{n,q}$ is applied to its
corresponding candidate representation $\mathbf{H}_{n,q,:}$.

\subsection{Effect of Additional LayerNorm in ER-LM}
\label{app:er_lm_layernorm}

As shown in Table~\ref{tab:er_lm_layernorm}, without $\mathcal{L}_{\mathrm{cwe}}$, adding LayerNorm to the LSTM output before projection improves occupancy IoU but reduces semantic mIoU. This may reflect a trade-off: normalizing memory-feature magnitudes can help distinguish occupied from empty regions while weakening magnitude cues useful for distinguishing semantic classes. With $\mathcal{L}_{\mathrm{cwe}}$, omitting this additional LayerNorm yields higher mIoU, TailIoU, and RegularIoU, at a small cost in ComplexIoU. We therefore omit this additional LayerNorm in our final configuration.

\begin{table}[htbp]
    \centering
    \caption{Effect of additional LayerNorm in ER-LM. Bold numbers indicate the best result in each column.}
    \label{tab:er_lm_layernorm}
    \scriptsize
    \setlength{\tabcolsep}{3.0pt}
    \renewcommand{\arraystretch}{1.14}
    \begin{adjustbox}{max width=\columnwidth}
    \begin{tabular}{@{}lccccc@{}}
        \toprule
        Method & IoU & mIoU & ComplexIoU & TailIoU & RegularIoU \\
        \midrule
        ER-LM & 36.09 & 12.53 & 22.96 & 1.32 & 11.57 \\
        ER-LM + LayerNorm & \textbf{36.53} & 12.45 & 23.05 & 1.26 & 11.26 \\
        ER-LM + LayerNorm + $\mathcal{L}_{\mathrm{cwe}}$ & 36.37 & 12.47 & \textbf{23.28} & 1.51 & 10.82 \\
        ER-LM + $\mathcal{L}_{\mathrm{cwe}}$ & 36.45 & \textbf{12.82} & 23.23 & \textbf{1.64} & \textbf{11.84} \\
        \bottomrule
    \end{tabular}
    \end{adjustbox}
\par\vspace{2pt}
\begin{minipage}{0.60\columnwidth}
\scriptsize\raggedright
\textit{Note.} All configurations were re-run under the same setting, which differs slightly from Table~\ref{tab:ablation}.
\end{minipage}
\end{table}

\subsection{Evaluation Protocol and Semantic Transition Metrics}
\label{app:semantic_transition_metrics}
Both models are evaluated on the same 815 SemanticKITTI validation frames using all nine decoder outputs. Predictions are evaluated at the ground-truth voxel resolution; invalid labels are excluded, and repeated samples introduced by distributed sampling are counted only once.

For a source layer $a$ and a target layer $b>a$, we compare semantic predictions at the same voxel locations. For each non-empty ground-truth class $c$, let $N_{c}^{\mathrm{CC}}$, $N_{c}^{\mathrm{CW}}$, $N_{c}^{\mathrm{WC}}$, and $N_{c}^{\mathrm{WW}}$ count correct-to-correct, correct-to-wrong, wrong-to-correct, and wrong-to-wrong transitions, respectively. For brevity, we omit the layer-pair indices $a\to b$ from these counts. Correctness requires an exact match to the ground-truth semantic label. Counts are pooled over the validation set before computing the class-wise degradation and correction rates:
\begin{align}
D_c^{a\to b}
&=\frac{N_c^{\mathrm{CW}}}{N_c^{\mathrm{CC}}+N_c^{\mathrm{CW}}},
\\
R_c^{a\to b}
&=\frac{N_c^{\mathrm{WC}}}{N_c^{\mathrm{WC}}+N_c^{\mathrm{WW}}}.
\end{align}
Degradation measures the fraction of initially correct predictions that become incorrect, whereas correction measures the fraction of initially incorrect predictions that become correct. To give each category equal weight, we define
\begin{align}
\mathrm{mD}^{a\to b}
&=\frac{100}{|\mathcal{C}_D^{a\to b}|}
\sum_{c\in\mathcal{C}_D^{a\to b}}D_c^{a\to b},
\\
\mathrm{mR}^{a\to b}
&=\frac{100}{|\mathcal{C}_R^{a\to b}|}
\sum_{c\in\mathcal{C}_R^{a\to b}}R_c^{a\to b}.
\end{align}
For each metric and layer pair, the class set contains only non-empty categories with a non-zero denominator in both models. Undefined rates are excluded rather than set to zero. Thus, each row compares the models over identical eligible classes, although the mD class set can vary across rows. Lower mD and higher mR indicate more favorable semantic preservation and correction, respectively.

\end{document}